\documentclass{article}
\usepackage{ijcai20}

\usepackage{times}
\usepackage{authblk}
\usepackage{soul}
\usepackage{url}
\usepackage[hidelinks]{hyperref}
\usepackage[utf8]{inputenc}
\usepackage[small]{caption}
\usepackage{epsfig}
\usepackage{graphicx}
\usepackage{amsmath}
\usepackage{amsthm}
\usepackage{booktabs}
\usepackage{algorithm}
\usepackage{algorithmic}
\usepackage{siunitx}
\usepackage{enumerate}
\usepackage[shortlabels, inline]{enumitem}
\usepackage{float}
\usepackage[numbers]{natbib}

\makeatletter
\newcommand{\printfnsymbol}[1]{%
  \textsuperscript{\@fnsymbol{#1}}%
}
\makeatother

\title{SimPropNet: Improved Similarity Propagation for Few-shot Image Segmentation}

\author[2]{Siddhartha Gairola\thanks{work done as part of Adobe MDSR internship program}}
\author[1]{Mayur Hemani\thanks{equal contribution}}
\author[1]{Ayush Chopra\printfnsymbol{2}}
\author[1]{Balaji Krishnamurthy}
\affil[1]{Media and Data Science Research Lab, Adobe Experience Cloud}
\affil[2]{IIIT Hyderabad}

\begin{document}
\maketitle

\begin{abstract}
Few-shot segmentation (FSS) methods perform image segmentation for a particular object class in a target (query) image, using a small set of (support) image-mask pairs. Recent deep neural network based FSS methods leverage high-dimensional feature similarity between the foreground features of the support images and the query image features. In this work, we demonstrate gaps in the utilization of this similarity information in existing methods, and present a framework - \textit{SimPropNet}, to bridge those gaps. We propose to jointly predict the support and query masks to force the support features to share characteristics with the query features. We also propose to utilize similarities in the background regions of the query and support images using a novel foreground-background attentive fusion mechanism. Our method achieves state-of-the-art results for one-shot and five-shot segmentation on the PASCAL-$5^i$ dataset. The paper includes detailed analysis and ablation studies for the proposed improvements and quantitative comparisons with contemporary methods. 
\end{abstract}
\vspace{-3mm}
\section{Introduction}
\label{secIntro}
Semantic image segmentation assigns class labels to image pixels. It finds applications in image editing \cite{SemanticSoftSeg, Tai, Tan}, medical diagnosis \cite{HyperDenseNetAH, UNet, MultiscaleGA}, automated driving \cite{Cordts_2016_CVPR} etc. Supervised deep neural network methods such as \cite{SegNet, dlv1, dlv2, DLV3, dlv3plus, DeepHR, psp} enable highly accurate image segmentation. However, they work for only a small number of fixed object classes, and require a large number of image-mask pairs for training which are hard to manually annotate. In several practical scenarios, including online commerce and design, the images may exist in a large number of sparsely populated classes (for instance, images of products). In such cases obtaining image-mask pairs for all possible classes to train a supervised method may be infeasible. Thus, segmentation methods that generalize to new classes with scant training data are of significance. Few-shot image segmentation methods, like \cite{OSLFSS, PANet, CANet}, are class-agnostic and alleviate the need for a large number of image-mask pairs. These methods utilize additional images and their masks of a particular class in predicting the binary segmentation mask for a given image of the same class. 
\begin{figure}[t]
\begin{center}
    \centering
    \includegraphics[width=0.98\linewidth]{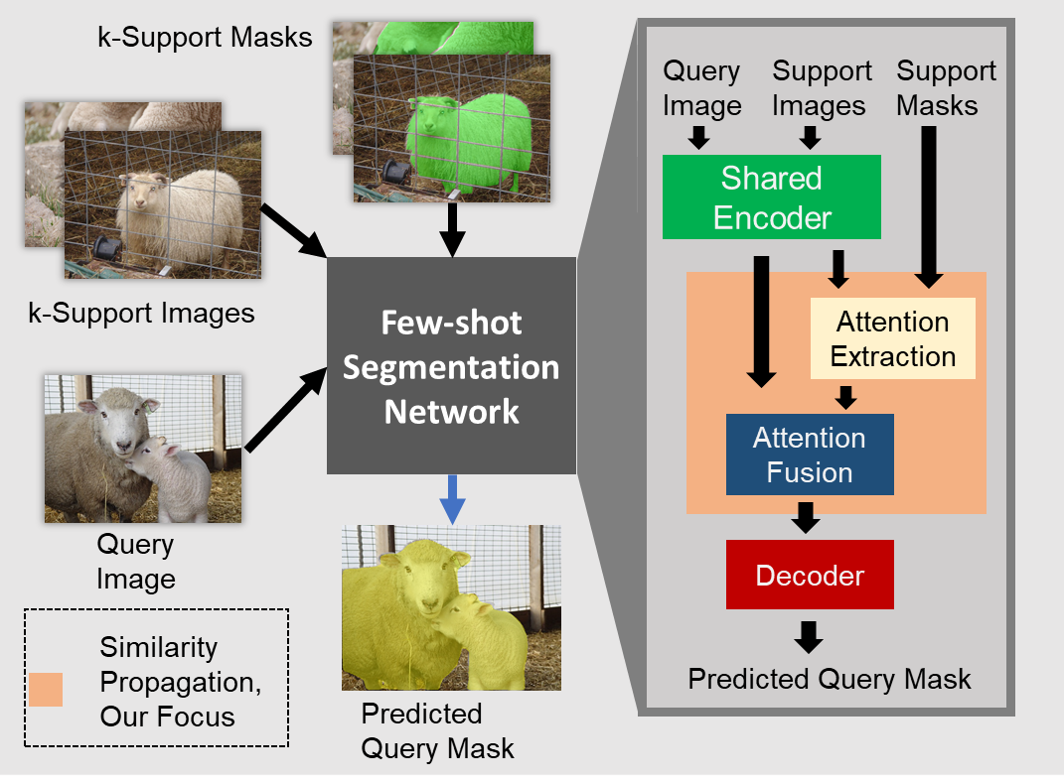}
\end{center}
\vspace{-4mm}
    \caption{Few-shot Image Segmentation: Broad architecture of contemporary methods (\cite{PANet, SGOne, CANet}). Features from the support images (in the support mask regions) are processed to obtain a probe representation and fused with features from the query image, and decoded to predict the query mask. Improving similarity propagation between the support and query branches is the focus of this work. The yellow mask in the diagram is a (1-shot) result from our method.\vspace{-1mm}}
    \label{fig:task_seg}
\end{figure}
This work proposes a new few-shot segmentation framework that seeks to alleviate a fundamental limitation in existing techniques and significantly improves upon the state-of-the-art.

Few-shot segmentation (FSS) methods typically take as input - a \textit{query} image for which the segmentation mask is required, a set of \textit{support} images, and their segmentation masks (\textit{support masks}). One-shot segmentation is the extreme setting where only a single support image-mask pair is available. Our method achieves state-of-the-art performance in both one-shot and few-shot settings (about 5\% and 4\% gain respectively).

Recent deep neural network based FSS methods (like \cite{AMP, PANet, SGOne, CANet}) employ variations of the following broad process (\autoref{fig:task_seg}): \begin{enumerate}[nosep]
\item Query and support image features are extracted using shared, pre-trained (on ImageNet \cite{ILSVRC15}) network layers.
\item A probe representation for regions of attention in the support image is obtained from the support features and mask(s).
\item The attention representation is fused with the query features. 
\item The fused features are decoded to obtain the query mask.
\end{enumerate}
The attention extraction and fusion modules leverage the high-dimensional feature similarity between the query and the support images to selectively decode the query features to output the segmentation mask. The focus of this work is to demonstrate gaps in the propagation of this similarity information in existing methods and to bridge those gaps.

Experiments with two state-of-the-art methods - \cite{AMP} and \cite{CANet} (see Section \ref{secPremise}) reveal that FSS methods make errors in visually similar regions, across image pair samples of identical classes. They perform poorly when support is identical to query inputs as well. These results indicate that the class and visual similarity information is not propagated optimally across the support and query branches . 

We predict the support mask from the support features to endow the features with specificity with respect to the target class, which in turn aids in similarity propagation between the support and query. We also leverage the similarities in the background regions of the support and query images through a background attention vector computed from the support features and inverse mask, and fuse it with the query features. Finally, to prevent the network from over-fitting on training class-conditional similarities, we employ an input channel averaging input augmentation for the query input. With these improvements we achieve state-of-the-art performance on PASCAL $5^i$ dataset for both one-shot and five-shot segmentation tasks. 

The contributions of this work can be summarized as follows:
\begin{enumerate}[nosep]
\item It highlights gaps in existing FSS methods in fully utilizing the similarity between the support and query images.
\item  It introduces \textit{SimPropNet}, a few-shot segmentation framework that bridges those gaps and achieves state-of-the-art performance on the PASCAL $5^i$ dataset in both 1-shot and 5-shot evaluation settings. The framework employs:
\begin{enumerate}[a., nosep]
    \item A dual-prediction scheme (DPr) where the query and support masks are jointly predicted using a shared decoder, which aids in similarity propagation between the query and support features.
    \item A novel foreground-background attentive fusion (FBAF) mechanism for utilizing cues from background feature similarities between the query and support images.
\end{enumerate}
\end{enumerate}

\noindent  The next section places the proposed method in the context of previous work related to the problem.



\section{Related Work}
\label{secRelatedWork}
Recent methods for few-shot semantic segmentation (FSS) build a framework for one-shot segmentation and subsequently construct methods to use the framework for k-shot segmentation ($k > 1$). The proposed work follows the same methodology. 

As described in Section \ref{secIntro}, most FSS methods employ a dual branched neural network model with a support branch and a query branch. This model was introduced by \citet{OSLFSS} where the support branch is conditioned on the support input to predict the weights of the last layer in the query branch which then predicts the query mask. 

\citet{CoFCN} improve upon \cite{OSLFSS} by employing a \textit{late fusion} strategy for the support mask and support feature maps to segment the query image. Late fusion adopts a parametric module in a two branch setting, which fuses information extracted from the support set with the features of the query image to produce the segmentation mask. \citet{Dong2018FewShotSS} combine the late fusion methodology with the idea of prototypical networks introduced in \cite{PNFSL} to learn a metric space where distances to \textit{prototypes} represent class-level similarity between images. Their method uses the prototypes as guidance to fix the query set segmentation rather than obtaining segmentation directly from metric learning. \citet{SGOne} take a different approach to late fusion. They introduce \textit{masked average pooling} (MAP) that pools support features of the regions of interest in the support images, and fuses them with the query features using vector cosine-similarity. This \textit{attention} map is then decoded to predict the query's segmentation maps. Building on \cite{Dong2018FewShotSS}, \citet{PANet} combine Prototypical Learning \cite{PNFSL}, and MAP \cite{SGOne} to incorporate the support information into the segmentation process. \citet{CANet} adopt a learnable method through an attention mechanism to fuse support information from multiple support examples along with iterative refinement. Their method also uses MAP, but instead of fusing its output with the query features using cosine-similarity, they concatenate it with the query features and then decode the output.

These methods for few-shot segmentation depend on the support set to produce the segmentation for the query image, but fail to fully utilize the support information efficiently. We present evidence for this gap in Section \ref{secPremise} and subsequently propose SimPropNet, a few-shot segmentation framework that seeks to bridge this gap and improve performance.
\section{Method}\label{secMethod}
\begin{figure*}[t]
\begin{center}
    \centering
    \includegraphics[width=\linewidth]{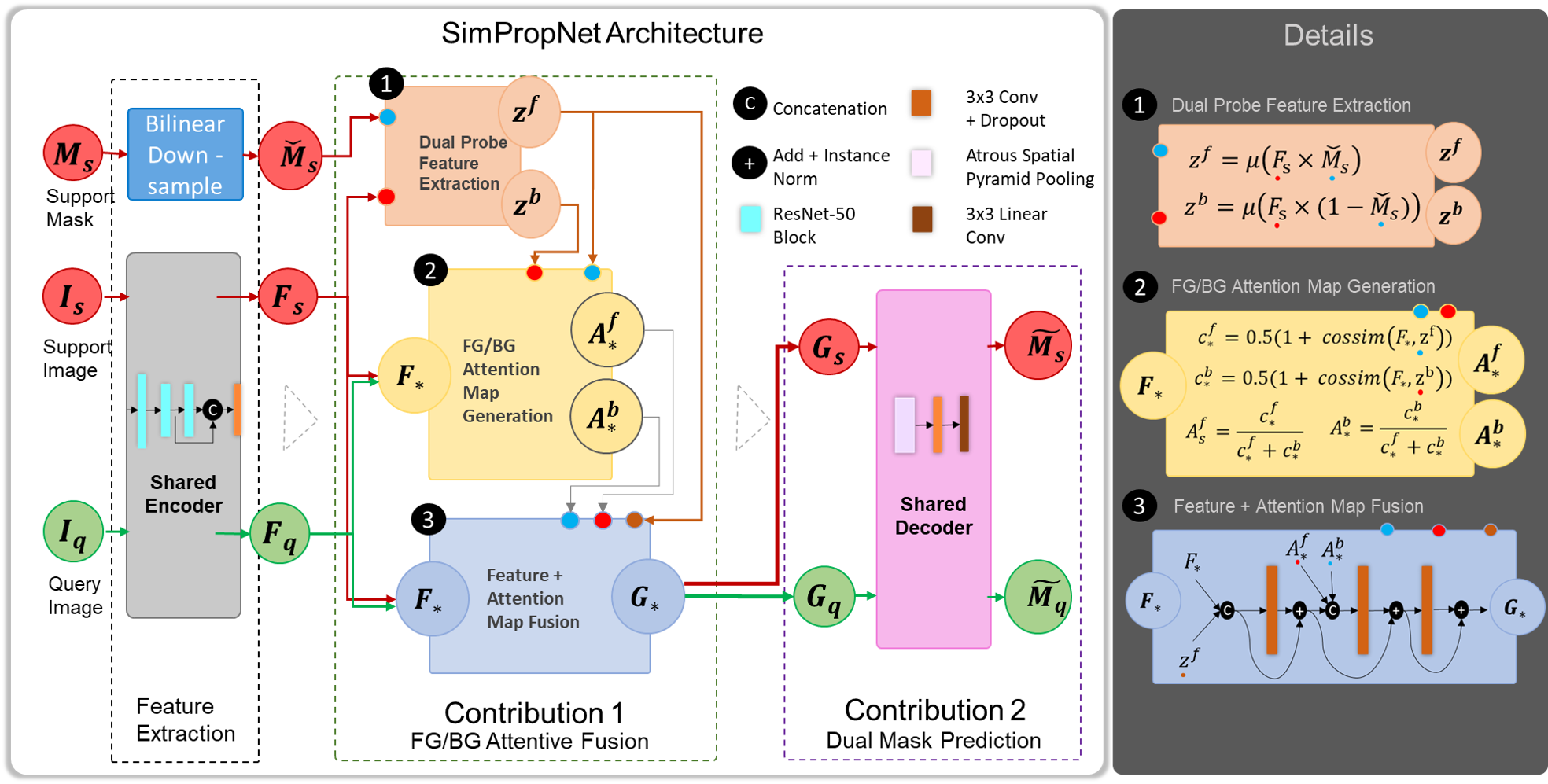}
\end{center}
\vspace{-4mm}
    \caption{SimPropNet Architecture: The support and query images are encoded to $F_s$, $F_q$ respectively, with pre-trained ResNet-50 layers and a single convolutional layer (as described in \cite{CANet}). The support features and the support mask are used to compute the foreground(FG) and background(BG) MAP vectors ($Z^f$, $Z^b$) where $\mu$ is average pooling. They are used to compute the four attention maps $A^f_s$,$A^b_s$, $A^f_q$, and $A^f_q$ (FG and BG for both support and query). The feature+attention fusion module combines ($F_s$, $Z^f$, $A^f_s$, and $A^b_s$) to obtain $G_s$, and identically, ($F_q$, $Z^f$, $A^f_q$, and $A^b_q$) to obtain $G_q$. These fused features ($G_s$, $G_q$) are finally decoded with the atrous spatial-pyramid pooling and convolutional layers to obtain the predicted masks $\tilde M_s$ and $\tilde M_q$ respectively.} 
    \label{fig:method}
\end{figure*}
In this section, we first establish firm ground to motivate our work, and then elucidate the proposed method in the one-shot segmentation setting. Subsequently, we demonstrate how we adapt the method for k-shot segmentation setting.

\subsection{Premise Validation}
\label{secPremise}
We first present experimental evidence validating the premise that there are gaps in similarity propagation between the support and the query images. This argument is expressed as follows:
\begin{enumerate}[nosep]
    \item FSS methods make errors for image regions that may not be inherently hard to segment.
    \item The regions of error in the support and query images are class-conditionally similar.
    \item Even with maximally similar support and query images, current FSS methods are unable to produce good predictions.
\end{enumerate}
We conduct experiments to corroborate this argument which indicates an opportunity to improve similarity sharing between the support and query images.

We also provide evidence to show that the background regions of the support and query images may be similar and could hold cues for improved segmentation of the query. The experiments use author-provided implementations of the recent state-of-the-art methods \cite{AMP} and \cite{CANet}. 

\paragraph{Errors by FSS versus Supervised Methods}
We measure the overlap (\autoref{tab:evidence_req}) between the error regions of output masks from two FSS methods (\cite{AMP} and \cite{CANet}) with the error regions of masks produced by DeepLab v3+ \cite{dlv3plus} (a state-of-the-art supervised method). The overlap between the error regions for FSS methods and DeepLab v3+ is very small, while the gap in the mean-IoU for correct predictions (DeepLab v3+ versus FSS methods) is large. This indicates that there are image regions where supervised methods like DeepLab v3+ do not fail (but FSS methods do), and that they are not characteristically difficult to segment. 

\begin{table}[]
    \centering
    \begin{tabular}{l|l|l|l}
    \toprule
    Method & FN Overlap & FP Overlap &  TP Gap \\
    \toprule
    CANet & 18.32 & 9.11 & 23.28 \\
    AMP & 10.99 & 19.06 & 39.41 \\
    \bottomrule
    \end{tabular}
    \vspace{-2mm}
    \caption{
    Percentage (\%) error overlap between Deeplab v3+ (DLv3+) \cite{dlv3plus}, and FSS Methods (CANet \cite{CANet}, and Adaptive Masked Proxies (AMP) \cite{AMP}). FN = False Negative (missed regions), FP = False Positives (incorrectly predicted as part of the mask). TP Gap = Overall gap in true prediction made by Deeplab v3+ and FSS methods. Low error overlap and high prediction gap indicates that FSS methods make different mistakes than DLv3+.}
    \label{tab:evidence_req}
\end{table}

\paragraph{Similarity of mispredicted regions}
To determine the similarity between regions of the query and support images where \cite{CANet} makes errors, we compute the Masked Average Pooling vectors (as described in \cite{SGOne}) using pre-trained VGG-16 \cite{VGGNet} features. This is done for several pairs of images of identical classes. for the following two regions of each image:
\begin{enumerate}[a., nosep]
    \item the pixels covered by the ground-truth masks ($Z^g$), and
    \item the pixels present in the regions mispredicted by the FSS method ($Z^e$).
\end{enumerate}
For each image pair $(A, B)$ of a class, the ratio of the inner-products of the MAP vectors for the error regions and the ground-truth mask regions - $(Z^e(A) \cdot Z^e(B))/Z^g(A)\cdot  Z^g(B)$ - is a measure of the similarity of the corresponding error regions relative to the similarity of the ground-truth mask regions. We measure the ratio over a 1000 pairs of images from the PASCAL VOC dataset \cite{pascal-voc-2012} and find the average to be 0.87 (std. dev. = 0.25). The high value of relative similarity of error regions substantiates the claim that errors are committed in regions of similarity that could have possibly been propagated from the support to the query.

\paragraph{Identical Inputs: (Support = Query)}
\autoref{tab:sameImage} reports results from a third experiment with \cite{AMP} and \cite{CANet} where the same image is given as input for both support and query (including the ground-truth mask for the support). This constitutes a basic test for similarity propagation in the network. Ideally, the network must produce the exact mask as provided in the support input, because the query and the support are as similar as possible. However, both methods (\cite{AMP} and \cite{CANet}) perform poorly for these inputs indicating the loss of similarity information in the networks.
\begin{table}[!htbp]
    \centering
    \begin{tabular}{p{0.15\linewidth}|p{0.15\linewidth}|p{0.15\linewidth}|p{0.15\linewidth}|p{0.15\linewidth}}
    \toprule
    Method & split-1 & split-2 & split-3 & split-4 \\
    \toprule
    CANet & 54.51  & 63.98 & 48.20 & 52.76 \\
    AMP & 54.41  & 69.34 & 64.79 & 60.02 \\
    \bottomrule
    \end{tabular}
    \vspace{-2mm}
    \caption{Percentage (\%) mean-IoU measured for FSS methods (CANet \cite{CANet} and AMP \cite{AMP}) for the test images from PASCAL $5^i$ dataset when query image ($I_{q}$) = support image ($I_{s}$).}
    \label{tab:sameImage}
\end{table}

\begin{figure}[!hbtp]
\begin{center}
    \centering
    \includegraphics[width=0.9\linewidth]{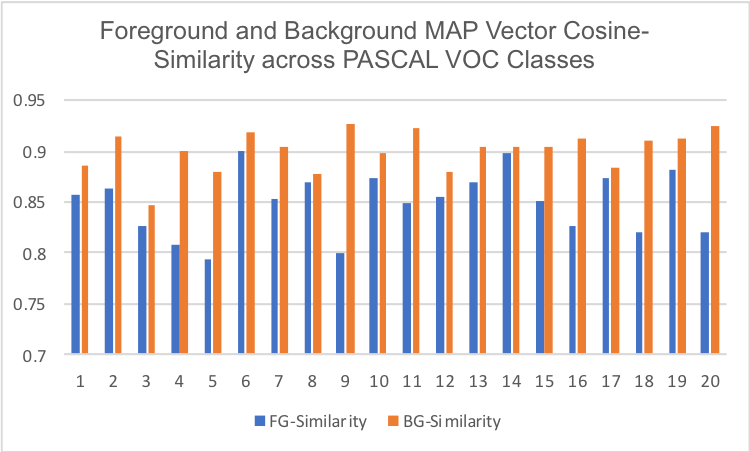}
\end{center}
\vspace{-4mm}
    \caption{Foreground and Background cosine-similarity based on MAP vectors computed using ResNet-50 (layer2+layer3) features for 4000 image pairs from the PASCAL VOC dataset. Background similarity has a greater average magnitude than foreground similarity across all classes. This indicates the opportunity to obtain cues from similar background regions.} 
    \label{fig:bgsimevidence}
\end{figure}

\paragraph{Cues from Background Similarity}
\autoref{fig:bgsimevidence} indicates the degree of similarity measured as the cosine-similarity between the MAP vectors of foreground and background regions for the pairs of images. The figure indicates that images of identical object classes have higher degree of feature similarity in their background features, than in the actual foreground features. This represents an opportunity to collect more information for accurate segmentation.  

The results of these experiments indicate the gaps in similarity propagation in existing FSS methods that we propose to exploit. Next, we describe the organization of the proposed network and how it addresses the similarity propagation problem. 

\begin{table*}[t]
\centering
\begin{tabular}{ c | c c c c c | c c c c c}
\toprule
Method & & & 1-shot & & & & & 5-shot & & \\

\cmidrule(lr){2-6}
\cmidrule(lr){7-11}

& split-1 & split-2 & split-3 & split-4	& mean & split-1 & split-2 & split-3 & split-4	& mean \\
\cmidrule(lr){1-1}
\cmidrule(lr){2-6}
\cmidrule(lr){7-11}
						
SimPropNet(ours) & \textbf{54.86}  & \textbf{67.33} & \textbf{54.52} & \textbf{52.02} & \textbf{57.19} & \textbf{57.20} & \textbf{68.50} & \textbf{58.40} & \textbf{56.05} & \textbf{60.04}  \\

CA-Net \cite{CANet} & 50.41 & 63.02 & 46.09 & 47.46 & 51.75 & 52.27 & 65.29 & 47.15 & 50.50 & 53.81  \\

PA-Net \cite{PANet} & 42.40 & 58.00 & 51.10 & 41.20 & 48.13 & 51.80 &   64.60 & 59.80 & 46.50 & 55.70 \\

SG-One \cite{SGOne} & 40.20 & 58.40 & 48.40 & 38.40 & 46.30 & 41.90 & 58.60 & 48.60 & 39.40 & 47.10 \\

AMP \cite{AMP} & 41.90 & 50.20 & 46.70 & 34.70 & 43.40 & 41.80 & 55.50 & 50.30 & 39.90 & 46.90 \\

co-FCN \cite{CoFCN} & 36.70 & 50.60 & 44.90 & 32.40 & 41.10 & 37.50 & 50.00 & 44.10 & 33.90 & 41.40 \\

OSLSM \cite{OSLFSS} & 33.60 & 55.30 & 40.90 & 33.50 & 40.80 &35.90 & 58.10 & 42.70 & 39.10 & 43.90 \\


 \bottomrule
\end{tabular}
\vspace{-2mm}
\caption{Percentage (\%) mean-IoU of one-shot segmentation on the PASCAL $5^i$ dataset using random partitions. The best results are highlighted in bold. SimProp-Net achieves state-of-the-art performance on both the 1-shot and 5-shot settings.}
\label{tab:ret_map}
\end{table*}

\subsection{Network Organization}
\autoref{fig:method} illustrates the architecture of SimPropNet. The network is organized  into two branches (support and query) with a shared encoder, a fusion module and a shared decoder. For our experiments, we use a ResNet-50 \cite{DBLP:journals/corr/HeZRS15} based feature extractor, and atrous spatial pyramid pooling based decoder as in \cite{CANet}. The encoder comprises of three layers from a ResNet-50 network pre-trained on ImageNet \cite{ILSVRC15}, and a single $3\times3$ dilated (rate = 2) convolutional layer with 256 filters. The ResNet-50 layers are frozen during training. The decoder comprises of an atrous spatial pyramid pooling layer (as introduced in \cite{dlv2}), and two convolutional layers. The last layer has linear activation and produces a 2-channel output $(1/8)^{th}$ the input size. The output from the last layer is resized to the expected mask size with bilinear interpolation. The predicted query and support masks are compared to their respective ground-truths using the cross-entropy loss.

\subsection{Support and Query Mask Prediction (DPr)}
The network is trained to predict both the support and the query masks using the same encoder and decoder. We submit that this dual prediction requirement forces the query and support features from the shared encoder to share greater and more target-specific similarity. For instance, if the support image (and mask) is of a car, the network must be able to recover back the support mask (i.e. the entirety of the car) from the foreground support features. Because the encoder is shared between the support and the query images, the query features share this characteristic with the support features. This is reflected in the effective gain in the mIoU as discussed in Section \ref{secAnalysis}.

\subsection{Foreground/Background Attentive Fusion (FBAF)}\label{secFBAF}
Late fusion methods (like \cite{ CoFCN, PANet, SGOne, CANet}) fuse the foreground features from the support branch into the query branch to locate the regions of class-conditional similarity. The foreground features are obtained using the masked average pooling (MAP) operation that computes a channel-wise weighted average of the support features where the weights are the support mask values at each position. We submit that fusing the background feature information from the support branch has the effect of suppressing similar background features in the query features. The fusion process can be concisely stated using the following three equations, each representing a step of the process:
\begin{enumerate}[nosep]
    \item Dual Probe Feature Extraction: The foreground and the background MAP vectors are computed using the support mask and the support features.
    \begin{align}\begin{split}
Z^f =& \mu_c(F_s * \Breve{M_s})\\
Z^b =& \mu_c(F_s * (1 - \Breve{M_s}))
\end{split}\end{align}
Here, $\mu_c$ is the average pooling operation with a kernel size equal to the feature map size, $F_s$ are the support features, and $\Breve{M_s}$ is the support mask down-sampled to the height and width of $F_s$.
    \item FG/BG Attention Map Generation: Four attention maps, two (foreground and background) each from the support and query features are computed.
\begin{align}\begin{split}
C(F, Z) =& (1 + cossim(F, Z))/2 \\ 
N(A, B) =&  (A/(A+B),\;\;B/(A+B)) \\
A^f_s,\;\;A^b_s =& N(C(F_s, Z^f),\;\;C(F_s, Z^b)) \\
A^f_q,\;\;A^b_q =& N(C(F_q, Z^f),\;\;C(F_q, Z^b))
\end{split}\end{align}
Here $cossim$ is the cosine-similarity measure and $F_q$ are the query features.
    \item Feature and Attention Map Fusion: The attention maps and the features for the query and support are fused to two feature vectors that are finally decoded into the query and support mask predictions respectively.
\begin{align}\begin{split}
G^0_* = & F_* \small{\oplus} Z^f \\
G^1_* = & IN(Conv_{3\times3}( G^0_* ) + G^0_*)\\
G^2_* = & IN(Conv_{3\times3}( G^1_* \oplus A^f_* \oplus A^b_* )+ G^1_*)\\
G_* = & IN(Conv_{3\times3}( G^2_* ) + G^2_*)\\
\end{split}\end{align}
Here $G_*$ is for both support ($G_s$) and query ($G_q$) features, and $\oplus$ is the concatenation operation and $IN$ is the Instance-Normalization operation \cite{InstanceNorm}.
\end{enumerate}

Because the fusion process combines information from both the foreground and the background support features, we term this \textit{FG/BG Attentive fusion} (FBAF). The analysis in Section \ref{secAnalysis} demonstrates the effective increase in prediction accuracy by employing FBAF. 


\subsection{k-Shot Inference}
The network is not specifically trained for k-shot training ($k>1$). To incorporate more than one support example pairs in inference, the MAP vectors computed in the Dual Probe Feature Extraction step are averaged over the support pairs:
\begin{equation}
    Z^f_{kshot} = \frac{\sum_{k}{Z^f_k}}{k},\;\;\;  Z^b_{kshot}= \frac{\sum_{k}{Z^b_k}}{k}
\end{equation}
These MAP vectors are used to compute the foreground and background attention maps, and are fused with the query features (Section \ref{secFBAF}) to predict the query segmentation mask. 

\subsection{Implementation Details}
The training is done on virtual machines on Amazon AWS with four Tesla V100 GPUs and 16-core Xeon E5 processors. The standard SGD optimization algorithm is used, learning rate is kept at (\num{2.5e-3}) and the batch size is 8 for all training. Training with still higher learning rate yields even better results than reported in the paper, but the training is not always stable and may decay considerably in later training steps. Training for each split is run for 180 epochs and the checkpoint with the highest validation score (mIoU) is retained. To prevent the network from overfitting on the training classes, we also use an input regularization technique called \textbf{Input Channel Averging} (\textbf{ICA}) where the query RGB image is mapped to a grayscale input (after normalizing) with a \textit{switch probability} (initialized at 0.25 for our experiments) that decays exponentially as training progresses. The particular benefit of using ICA is discussed in Section \ref{secAnalysis}.


\vspace{-4mm}
\section{Results}\label{secResults}
\begin{figure}[t]
\begin{center}
    \centering
    \includegraphics[width=1.05\linewidth]{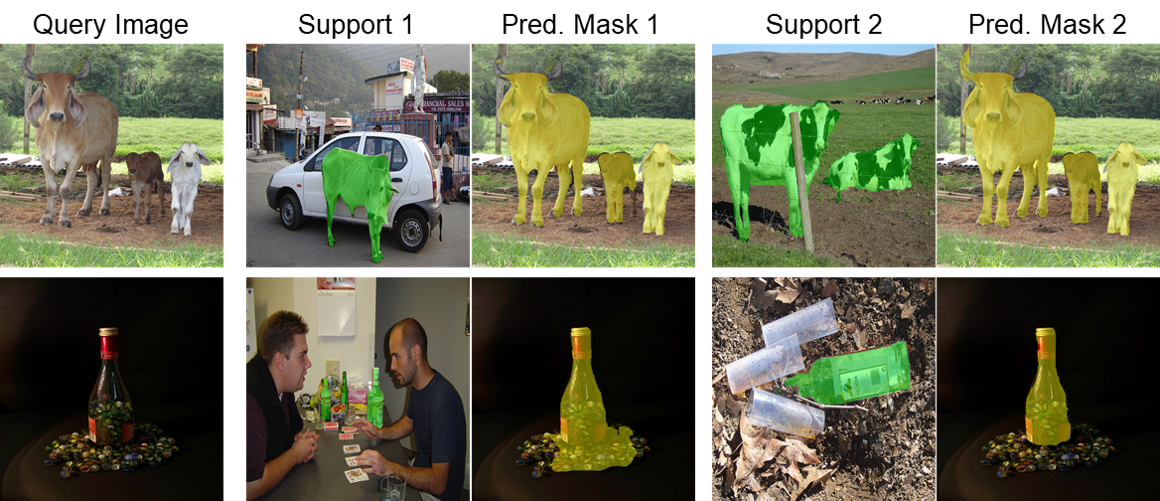}
\end{center}
\vspace{-4mm}
    \caption{Qualitative One-shot Segmentation Results from SimPropNet. Notice that providing a more similar support image helps to improve the segmentation (Cow's horn in the top row, and marbles in the bottom row).} 
    \label{fig:qualresults1}
\end{figure}

\begin{figure}[t]
\begin{center}
    \centering
    \includegraphics[width=1.05\linewidth]{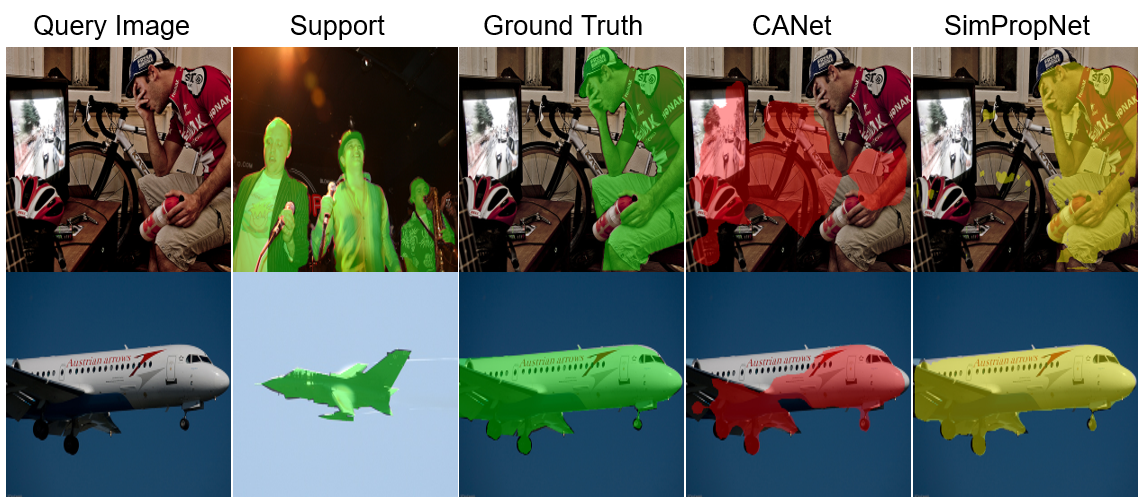}
\end{center}
\vspace{-4mm}
    \caption{One-shot segmentation results compared to CANet\cite{CANet} output. Row 1 indicates improved class-specificity - the mask is localized on the correct target class. Row 2 demonstrates better utilization of similar background context.} 
    \label{fig:qualresults2}
\end{figure}
In this section, we present the evaluation metrics and reporting criteria for few-shot segmentation, the quantitative results for SimPropNet, and comparison with other state-of-the-art methods. For brevity, we include very few qualitative results highlighting the key benefits our work. 

\subsection{Metric and Reporting}
For consistency of comparison with baseline techniques, we follow the evaluation protocol established by \citet{OSLFSS}. Different instances of the network are trained with the different training splits of the PASCAL-$5^i$ dataset, each with 15 classes of objects and their masks. Testing is done on the images with objects of the 5 withheld classes. The mean intersection-over-union (mIoU) metric for output masks with respect to the ground-truth mask is computed. We report the mean IoU values for random support and query pairs (images of the same class) in the standard 1-shot and 5-shot ($k = 5$ support images) settings. To ensure fairness in comparison, we use author-provided code for baselines and evaluate performance for three sets of random support-query test pairs for each of the splits and report the average performance. These splits will be released along with the paper upon acceptance.
\subsection{SimPropNet Results}
\label{secPartitions}
\paragraph{Quantitative Results}
We compare our performance using the method described in \cite{OSLFSS} that employs random query and support image pairs (partitions) from each test class in PASCAL-$5^i$ splits. The mean intersection-over-union (mIoU) values are reported in Table \ref{tab:ret_map}.
Our method, SimPropNet, achieves state-of-the-art performance in both the 1-shot and the 5-shot setting. SimPropNet outperforms CANet \cite{CANet}, the current state-of-the-art in 1-shot setting, by 5.44\% in 1-shot and 6.23\% in the 5-shot evaluation. Further, SimPropNet outperforms PANet \cite{PANet}, the current state-of-the-art in 5-shot setting, by 9.06\% in the 1-shot evaluation and 4.34\% in the 5-shot evaluation task. 

\begin{table}[!htbp]
\centering
\begin{tabular}{ c | c c}
\toprule
Method & 1-shot & 5-shot \\
\cmidrule(lr){1-3}
SimPropNet(ours) & \textbf{72.96}  & \textbf{ 72.90}\\
FWB \cite{FWB} & - & - \\
PGN \cite{PGN19} & 69.9 & 70.5 \\
PA-Net \cite{PANet} & 66.5 & 70.7 \\
CA-Net \cite{CANet} & 66.2 & 69.6 \\
SG-One \cite{SGOne} & 63.1 & 65.9\\
AMP \cite{AMP} & 62.2 & 63.8\\
OSLSM \cite{OSLFSS} & 61.3 & 61.5\\
co-FCN \cite{CoFCN} & 60.1 & 60.2\\
 \bottomrule
\end{tabular}
\vspace{-2mm}
\caption{Comparison tables with the FG-BG metric as used in \cite{CoFCN} for the 1-shot and 5-shot setting.}
\label{tab:fg_bg}
\end{table}

\paragraph{Qualitative Results}
Qualitative one-shot segmentation results also indicate significant improvements in output. Figure \ref{fig:qualresults1} highlights how providing a more similar support image and mask helps in improving the segmentation, and is expected in a practical scenario. Figure \ref{fig:qualresults2} presents two sample results for comparison with CANet \cite{CANet}. The first row illustrates how CANet may overfit on training classes (target test class (person) is from split-2, and bicycle is in the training set) or its features lack specificity to the target class, and how SimPropNet overcomes this issue. Row 2 of the figure illustrates the benefit of providing support images with similar backgrounds.

\section{Analysis}
\label{secAnalysisDiscussion}
In this section, we first analyse the particular benefit of each component contribution of SimPropNet individually. Next, we probe the effectiveness of SimPropNet in improving similarity propagation. We present these evaluations in the one-shot setting and compare against CANet \cite{CANet}, the existing state-of-the-art.
\subsection{Ablation Study of Components}
\label{secAnalysis}
We study the effectiveness of each of our contributions individually - dual prediction (DPr) of the support and query masks, and foreground-background attentive fusion (FBAF). \autoref{tab:ablation_components} reports the mIoU values over the different splits of the PASCAL $5^i$ dataset. As highlighted by the results, both DPr and FBAF used individually demonstrate clear gains over the baseline (CANet \cite{CANet}) of 4.14\% and 3.75\% in mIoU respectively. FBAF alone performs better than DPr in three of the four splits, but has a slightly worse mean because of its sharp decline in performance in split-3. The combination of DPr and FBAF achieves an improvement of 5.34\% over \cite{CANet}. Additionally using ICA during training further improves mIoU on three of the splits and increases mean mIoU to 57.19\%.

\begin{table}[!htbp]
    \centering
    \begin{tabular}{p{0.22\linewidth}|p{0.11\linewidth}|p{0.11\linewidth}|p{0.11\linewidth}|p{0.11\linewidth}|p{0.09\linewidth}}
    \toprule
    Method & split-1 & split-2 & split-3 & split-4 & mean\\
    \toprule
    		
	Baseline (CANet)& 50.41 & 63.02 & 46.09 & 47.46  & 51.75\\
	\midrule
    DPr & 52.69  & 66.57 & 53.1 & 51.23 & 55.89 \\
    FBAF & 54.16  & 66.71 & 49.11 & 52.00  & 55.50 \\
    DPr + FBAF  & 54.08  & 67.29 & 54.05 & \textbf{52.93} & 57.09\\
    \midrule
    SimPropNet &\textbf{54.86} & \textbf{67.33} & \textbf{54.52} & 52.02 & \textbf{57.19}\\
    \bottomrule
    \end{tabular}
    \vspace{-2mm}
    \caption{Ablations for the different components of SimPropNet (DPr + FBAF + ICA). DPr is the joint prediction of query and support masks, FBAF is the FG/BG Attentive Fusion module, and ICA is the input channel averaging regularization.}
    \label{tab:ablation_components}
\end{table}

\subsection{Measuring Gain in Similarity Propagation}
We evaluate the performance of the proposed network on identical support and query images (as reported for \cite{CANet} and \cite{AMP} in \autoref{tab:sameImage} in Section \ref{secPremise}). The performance of a one-shot segmentation method on identical inputs for the query and the support is the upper bound on the degree of similarity propagation in the network. Results of this experiment for SimPropNet (\autoref{tab:sameImage_before_after}) show an average gain of 21.5\% over CANet \cite{CANet}, and 14.23\% over AMP \cite{AMP}, over all splits. This indicates clearly that the network is utilizing the similarity information between the query and support images better.

\begin{table}[!htbp]
    \centering
    \begin{tabular}{p{0.18\linewidth}|p{0.11\linewidth}|p{0.11\linewidth}|p{0.11\linewidth}|p{0.11\linewidth}|p{0.11\linewidth}}
    \toprule
    Method & split-1 & split-2 & split-3 & split-4 & mean \\
    \toprule
    SimPropNet & 71.24 & 82.09 & 74.97 & 77.15 & 76.36 \\
    \midrule
    CANet & 54.51 &	63.98 &	48.2 &	52.76  & 54.86\\
    $\Delta$CANet & \textbf{16.73} & \textbf{18.11} & \textbf{26.77} & \textbf{24.39} & \textbf{21.50} \\
    \midrule
    AMP & 54.41 & 69.34 &	64.79 &	60.02  & 62.14 \\
    $\Delta$AMP &  \textbf{16.83} & \textbf{12.75} & \textbf{10.18} & \textbf{17.13} & \textbf{14.23} \\
    \bottomrule
    \end{tabular}
    \vspace{-2mm}
    \caption{Percentage (\%) mean-IoU measured for FSS methods as compared with the proposed SimPropNet for test images from PASCAL $5^i$ dataset when query image ($I_{q}$) = support image ($I_{s}$) }
    \label{tab:sameImage_before_after}
\end{table}

\section{Conclusions and Future Work}

The paper presents a rigorous argument that similarity propagation in existing few-shot image segmentation networks is sub-optimal. It proposes SimPropNet, a deep neural network with two strategies for bridging this gap - predicting the support mask besides the query mask with a shared decoder, and fusing background features into the feature fusion module. The network achieves a new state-of-the-art as shown by a comprehensive set of experiments. 
Class-conditional similarity matching can only match pixels with a similar class-mix between the query and the support images. In future work, we focus on exploring the "objectness" aspect of the target image to improve few-shot segmentation.

\bibliographystyle{named}


\begin{figure*}[!htbp]
\begin{center}
    \centering
    \includegraphics[width=0.8\linewidth]{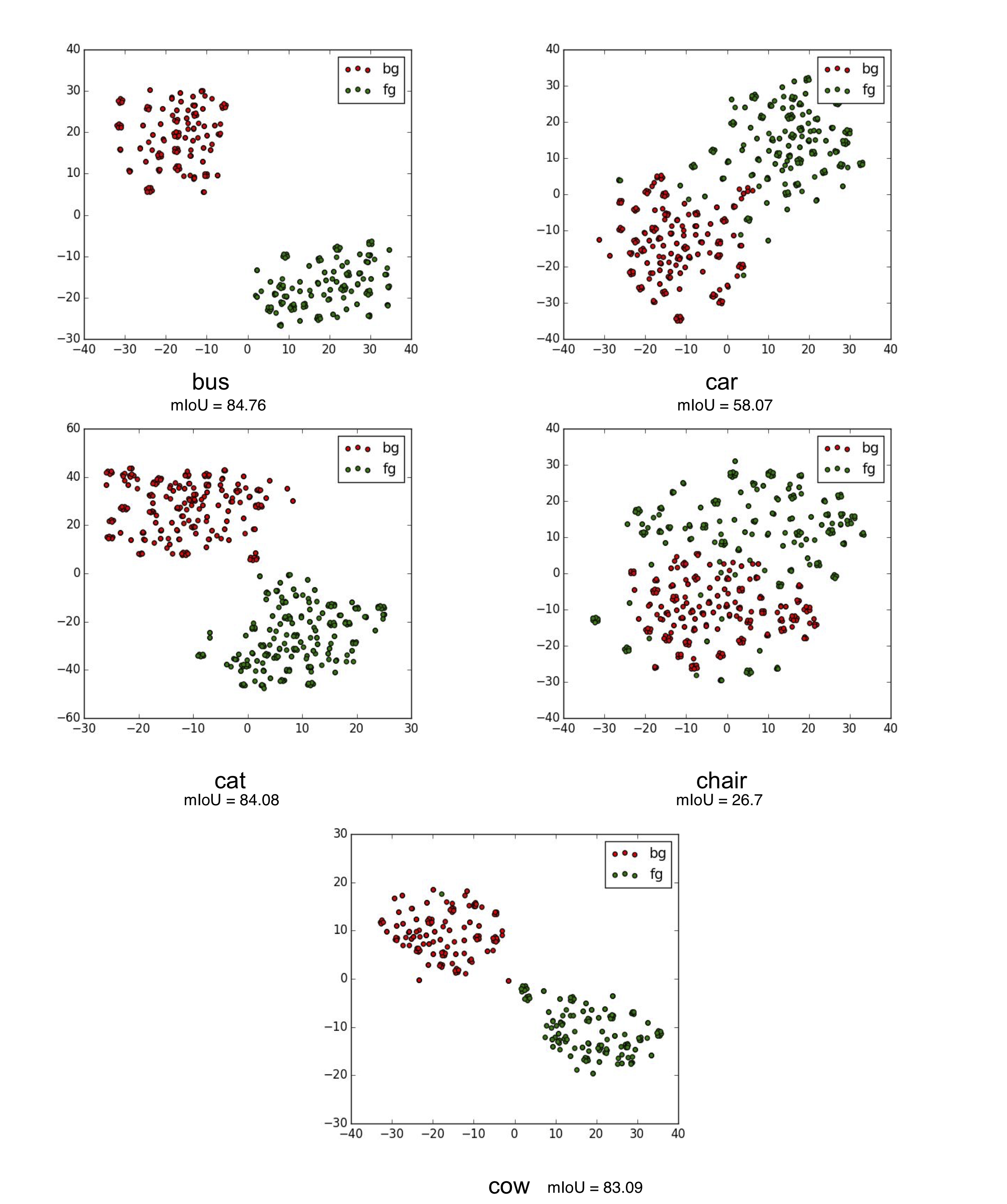}
\end{center}
\vspace{-4mm}
    \caption{tSNE feature visualizations (for classes in split-1) for MAP vectors computed for foreground and background form neat clusters.}

    \label{fig:tsne}
\end{figure*}

\end{document}